\documentclass{article} 
\usepackage{arxiv,times}

\usepackage{amsmath,amsfonts,bm}

\def\eqref#1{equation~\ref{#1}}

\def\1{\bm{1}}

\DeclareMathAlphabet{\mathsfit}{\encodingdefault}{\sfdefault}{m}{sl}
\SetMathAlphabet{\mathsfit}{bold}{\encodingdefault}{\sfdefault}{bx}{n}

\usepackage{url}
\usepackage{graphicx}
\usepackage{wrapfig}
\usepackage{booktabs}
\usepackage{multirow}
\usepackage{xcolor}
\usepackage{colortbl}
\usepackage[normalem]{ulem}
\definecolor{lightgray}{rgb}{0.9, 0.9, 0.9}

\definecolor{citecolor}{RGB}{34, 149, 34}
\usepackage[pagebackref=true,breaklinks=true,letterpaper=true,colorlinks,citecolor=citecolor,bookmarks=false]{hyperref}

\title{One-for-All: Generalized LoRA for \\ Parameter-Efficient Fine-tuning}

\author{Arnav Chavan$^{*1,2}$, Zhuang Liu$^3$, Deepak Gupta$^2$, Eric Xing$^{1,4}$, Zhiqiang Shen\thanks{Equal contribution. Project page: \url{https://sites.google.com/view/generalized-lora}.} ~$^1$ \\
$^1$MBZUAI $^2$Transmute AI Lab $^3$Meta AI Research $^4$CMU\\
\texttt{\{arnav.chavan,eric.xing,zhiqiang.shen\}@mbzuai.ac.ae} 
}

\newtheorem{theorem}{Theorem}

\iclrfinalcopy 
\begin{document}

\maketitle

\begin{abstract}
We present Generalized LoRA (GLoRA), an advanced approach for universal parameter-efficient fine-tuning tasks. Enhancing Low-Rank Adaptation (LoRA), GLoRA employs a generalized prompt module to optimize pre-trained model weights and adjust intermediate activations, providing more flexibility and capability across diverse tasks and datasets. Moreover, GLoRA facilitates efficient parameter adaptation by employing a scalable, modular, layer-wise structure search that learns individual adapter of each layer. Originating from a unified mathematical formulation, GLoRA exhibits strong transfer learning, few-shot learning and domain generalization abilities, as it adapts to new tasks through not only weights but also additional dimensions like activations. Comprehensive experiments demonstrate that GLoRA outperforms all previous methods in natural, specialized, and structured vision benchmarks, achieving superior accuracy with fewer parameters and computations. The proposed method on LLaMA-1 and LLaMA-2 also show considerable enhancements compared to the original LoRA in the language domain. Furthermore, our structural re-parameterization design ensures that GLoRA incurs no extra inference cost, rendering it a practical solution for resource-limited applications. 
Code and models are available at: \href{https://github.com/Arnav0400/ViT-Slim/tree/master/GLoRA}{GitHub}.
\end{abstract}

\section{Introduction}

Large-scale deep neural networks have revolutionized the field of artificial intelligence, demonstrating unprecedented performance across various tasks and domains. These highly complex models, often with millions or even billions of parameters, have demonstrated remarkable capabilities in areas such as computer vision~\citep{dosovitskiyimage}, natural language understanding~\citep{vaswani2017attention}, and speech recognition~\citep{radford2022robust}. Typically, these colossal models are pre-trained on general and large-scale datasets, such as ImageNet~\citep{5206848} or Web Crawl Text~\citep{wenzek2019ccnet}, and are subsequently adapted to downstream target scenarios through fine-tuning or transfer learning. Given the immense computational resources required by large pre-trained architectures, many parameter-efficient fine-tuning (PEFT) methods~\citep{hu2021lora,shen2021partial,jia2022visual,zhang2022neural,luo2023towards} have been proposed. For instance, Low-Rank Adaptation (LoRA)~\citep{hu2021lora} aims to reduce the number of trainable parameters by exclusively learning pairs of rank-decomposition matrices whilst keeping the original model parameter static. Adapter~\citep{houlsby2019parameter} implements bottleneck adapter modules and incorporates a modest number of task-specific parameters into a fixed pre-trained model. Similarly, Visual Prompt Tuning (VPT)~\citep{jia2022visual} introduces a minimal number of learnable parameters to the input of the Transformer, leaving the entire backbone frozen during fine-tuning.

However, distinct downstream datasets often possess unique characteristics, such as natural, specialized, and structured data, which differ significantly in distribution and composition. A static fine-tuning strategy may not sufficiently account for these disparities, thereby hindering its capacity to adapt to diverse datasets. To rectify this, we propose a flexible, parameter-efficient fine-tuning scheme in this work to manage the variations of multiple downstream datasets within a consolidated formulation. Our approach presents a generalized version of LoRA from a unified parameter-efficient fine-tuning perspective, amplifying LoRA's capability, scalability, and adaptability by rescaling and shifting intermediate activations, in conjunction with implementing a structural re-parameterization design, etc. It is challenging to devise a unified approach that integrates all adjustable dimensions and possibilities when tuning a pre-trained network, especially in the case of transformer architectures which contains various distinct modules, while our proposed approach presents a practicable solution to navigate this complexity.

Specifically, our approach presents a unified framework that can achieve comprehensive fine-tuning paradigms from a single formulation, i.e., a {\em One-for-All} fine-tuning architecture. It comprises a supernet, which, when optimized cost-effectively through evolutionary search, yields results that surpass those of prevailing fine-tuning methodologies necessitating expensive data-dependent hyperparameter search. The proposed approach exhibits the following advantages: ({1}) It concurrently takes into account multiple dimensions to enhance capability and flexibility during fine-tuning, encompassing weights, features, and input tokens. 
({2})  It conducts an implicit search devoid of any manual hyperparameter tuning, thus justifying the increased training time. ({3})  It incurs no additional inference cost thanks to our structural re-parameterization architecture, whereby the extra fine-tuning parameters will be fused to the proximate projection weights post-training.

We conduct comprehensive experiments on VTAB-1K~\citep{zhai2020the}, ImageNet~\citep{5206848} and its variants~\citep{recht2019imagenet,wang2019learning,hendrycks2021natural,hendrycks2021many}, and Huggingface leaderboard benchmarks~\citep{open-llm-leaderboard} for evaluating on language domain. The VTAB-1K dataset comprises 19 heterogeneous vision datasets, enveloping a broad spectrum of visual domains that include natural objects and scenes, textures and shapes, satellite imagery, among others. GLoRA surpasses all previous state-of-the-art PEFT methods by a substantial margin in terms of average accuracy. Additionally, we evaluate the model's few-shot learning capacity on five fine-grained visual recognition datasets, akin to prior works~\citep{zhang2022neural,jia2022visual}, along with its ability for domain generalization and robustness on ImageNet-V2~\citep{recht2019imagenet}, ImageNet-Sketch~\citep{wang2019learning}, ImageNet-A~\citep{hendrycks2021natural}, and ImageNet-R~\citep{hendrycks2021many} datasets. GLoRA significantly outperforms previous methods across all these benchmarks, without incurring any extra computational overhead during the inference phase. 

Our contributions:
\vspace{-0.05in}
\begin{itemize}
\addtolength{\itemsep}{-0.0in}
\item {We propose Generalized LoRA (GLoRA), a novel parameter-efficient fine-tuning framework. GLoRA enhances the low-rank adaptation approach with a more generalized prompt module design per layer, offering enhanced capability and flexibility in finetuning.}
\item {GLoRA presents a unified framework that achieves universal fine-tuning paradigms from a single formulation, i.e., a {\em One-for-All} \footnote{{\em One-for-All} represents that one formulation can be transformed into various shapes of PEFT paradigms.} fine-tuning architecture. During inference, the adapters yielded through GLoRA seamlessly integrate into the base network, resulting in no additional model weights. Thus, it incurs no extra inference computational load.}
\item {We conduct extensive experiments on large vision (ViT-B) and language models (LLaMA-1 and 2) with downstream fine-tuning, few-shot learning, and domain generalization using various datasets. Our experimental results demonstrate that GLoRA outperforms all previous methods on these benchmarks while requiring only a small number of extra tunable parameters in training and no additional inference cost.}
\end{itemize}

\section{GLoRA}

In this section, we start from providing a mathematical overview of existing state-of-the-art PEFT methods and discuss the advantages and disadvantages for them. Then, we introduce a unified formulation of integrating all existing state-of-the-art PEFT methods and elaborate our proposed generalized LoRA in detail following this unified formulation perspective. After that, a structural re-parameterization design is presented to show the inference efficiency without additional cost. An evolutionary search for optimal layer-wise configurations is also introduced to achieve the goal of generalized LoRA. We further give the theoretical analysis and discussion on the higher capability of the proposed method.

\subsection{Previous Solutions with Limitations}

\noindent{\bf Visual Prompt Tuning}~\citep{jia2022visual}: VPT introduces a small amount of task-specific learnable parameters into the input space while freezing the entire pre-trained Transformer backbone during downstream fine-tuning. It proposes two strategies: VPT-Shallow, where only input space has the trainable prompt:
\begin{equation}
\begin{array}{l} {\left[\mathbf{x}_{1}, \mathbf{Z}_{1}, \mathbf{E}_{1}\right] }=L_{1}\left(\left[\mathbf{x}_{0}, \mathbf{P}, \mathbf{E}_{0}\right]\right) \\ {\left[\mathbf{x}_{i}, \mathbf{Z}_{i}, \mathbf{E}_{i}\right] }=L_{i}\left(\left[\mathbf{x}_{i-1}, \mathbf{Z}_{i-1}, \mathbf{E}_{i-1}\right]\right) \end{array}
\end{equation}  
where $P$ is a trainable prompt. $\mathbf{x}$ is the [CLS] token, $\mathbf E$ are the image patches. Prompts use $<$1\% trainable parameters as compared to the original model.

VPT-Deep, where every layer has the trainable prompt. The formulation is:
\begin{equation}
\begin{array}{l} {\left[\mathbf{x}_{i}, \ldots, \mathbf{E}_{i}\right] }=L_{i}\left(\left[\mathbf{x}_{i-1}, \mathbf{P}_{i-1}, \mathbf{E}_{i-1}\right]\right) \end{array}
\end{equation}  
VTP-Deep outperforms full fine-tuning on many vision tasks and also has better accuracy in a low data regime. However, VPT increases cost in the inference stage which is not negligible. 

\noindent{\bf AdaptFormer}~\citep{chen2022adaptformer}: AdaptFormer introduces a parallel learnable branch of two linear layers and ReLU over the MLP block, and updates only this path while freezing other parts.
\begin{equation}
\tilde{x}_{\ell}=\mathbf{ReLU}\left(\mathrm{LN}\left(x_{\ell}^{\prime}\right) \cdot \mathbf{W}_{\mathrm{down }}\right) \cdot \mathbf{W}_{\mathrm{up }}
\end{equation}
\begin{equation}
x_{\ell}=\mathbf{MLP}\left(\mathrm{LN}\left(x_{\ell}^{\prime}\right)\right)+s \cdot \tilde{x}_{\ell}+x_{\ell}^{\prime}
\end{equation}
where $x_{\ell}^{\prime}$ are the tokens after MHSA at the $\ell$-th layer. $\mathbf{W}_{\mathrm{down }}$ and $\mathbf{W}_{\mathrm{up}}$ are weights corresponding to a down-projection layer and an up-projection layer from the parallel branch, respectively. $s$ is a scale factor. AdaptFormer also increases the inference cost due to the presence of a parallel branch.

\noindent{\bf LoRA}~\citep{hu2021lora}: LoRA proposes to freeze the pre-trained model weights and injects trainable low-rank decomposition matrices into each layer. It learns only the residual from pre-trained weight. Assuming $\mathbf{W}_0$, $\mathbf{b}_0$, $x$ are pre-trained weights, bias and input, let $f$ be a linear layer, thus $f(x) = \mathbf{W}_0x+\mathbf{b}_0$. During fine-tuning, $\mathbf{W}_0$ and $\mathbf{b}_0$ are frozen, the learning process will be:
\begin{equation}
\begin{array}{ll} 
 f(x)=\mathbf{W}_{0} x+\Delta \mathbf{W} x+\mathbf{b}_{0}=\mathbf{W}_\mathrm{LoRA} x+\mathbf{b}_{0} 
\end{array}
\end{equation}
where $\Delta \mathbf{W}$ is the low-rank decomposition weights that are learnable.

\noindent{\bf Scaling \& Shifting Features (SSF)}~\citep{lian2022scaling}: SSF module scales and shifts features after every MLP, MHSA, Layernorm module during training, and performs re-parameterization during inference as it is a linear structure.
\begin{equation}
\bm y=\bm \gamma \odot x+\bm \beta
\end{equation}
where $\bm y$ is the output features. $\bm \gamma$ and $\bm \beta$ are the scale and shift factors, $\odot$ is the dot product. 
This method has no increase in inference but the capability is limited to feature adaptation. 

\noindent{\bf FacT}~\citep{jie2022fact}: FacT proposes to use a tensorization-decomposition method to store the additional weight, the weights of the model are tensorized into a single 3D tensor, and their additions are then decomposed into lightweight factors. In fine-tuning, only the factors will be updated and stored.
\begin{equation}
\begin{array}{ll} & f(x)=\mathbf{W}_{0} x+\mathbf{b}_{0}+\mathbf{U} \Sigma \mathbf{V} x=\left(\mathbf{W}_{0}+\mathbf{U} \Sigma \mathbf{V}\right) x+\mathbf{b}_{0} 
\end{array}
\end{equation}
where $\Delta \mathbf{W}$ in LoRA is decomposed into $\mathbf{U}$, $\mathbf{V}$ and $\bm \Sigma$. This is {\em Tensor-Train} in FacT.
\begin{equation}
\begin{array}{ll}  f(x)=\mathbf{W}_{0} x+\mathbf{b}_{0}+\mathbf{U} \mathbf{C} \mathbf{P} \mathbf{V} x=\left(\mathbf{W}_{0}+\mathbf{U} \mathbf{C} \mathbf{P} \mathbf{V}\right) x+\mathbf{b}_{0} 
\end{array}
\end{equation}
where $\Delta \mathbf{W}$ in LoRA is decomposed into $\mathbf{U}$, $\mathbf{C}$, $\mathbf{P}$ and $\mathbf{V}$. This is {\em Tucker }  in FacT.

\noindent{\bf RepAdapter}~\citep{luo2023towards}: 
RepAdapter inserts lightweight networks into the pre-trained models, and the additional parameters will be re-parameterized to the nearby projection weights after training.
Adding sequential (not parallel) adapter to both MHSA and MLP, adapter is linear thus allowing for re-parameterization. It contains two layers: downsampling dense FC layer to downsample inputs; upsampling downsampled features that are divided into groups, and each group has an upsampling layer. The group of upsampling layers can be merged into a single sparse upsampling layer and can be re-parameterized directly into the original MLP/MHSA. The formulation can be:
\begin{equation}
\begin{array}{ll}  f(x)&=\mathbf{W}_{0}\left(x+\mathbf{W}_{u}\left(\mathbf{W}_{d} x+\mathbf{b}_{d}\right)+\mathbf{b}_{u}\right)+\mathbf{b}_{0}\\&=\left(\mathbf{W}_{0}+\mathbf{W}_{0} \mathbf{W}_{u} \mathbf{W}_{d}\right) x+\mathbf{W}_{0} \mathbf{W}_{u} \mathbf{b}_{d}+\mathbf{W}_{0} \mathbf{b}_{u}+\mathbf{b}_{0}\end{array}
\end{equation}
where $\mathbf{W}_u$, $\mathbf{W}_d$, $\mathbf{b}_u$ and $\mathbf{b}_b$ are learnable weights and biases, respectively.

\noindent{\bf Limitations:} In general, many existing PEFT methods such as (VPT, Adapter) increase the inference time since the proposed structure cannot be re-parameterized. Direct prompt tuning is also hard to design as it brings in computational burden and requires hyper-parameter tuning i.e., how and where to place prompts. LoRA can be re-parameterized at inference but it does not scale up for larger matrices and the adaptation ability is constrained on weight space. SSF / Repadaptor cannot learn the wieght change i.e., $\Delta \mathbf{W}$ in weight space, whereas LoRA / FacT cannot efficiently learn the scaling and shifting of feature change i.e., $\Delta \mathbf{H}$ in features space. Both feature and weight spaces need flexibility while performing transfer learning from a large model. 
Our proposed idea in this work attempts at: $\Delta \mathbf{W}$ tuning, $\Delta \mathbf{H}$ tuning, along with $\mathbf{W}$ and $\mathbf{H}$ scale and shift learning.

\subsection{A Unified Formulation of One-for-All} \label{method}

\begin{wrapfigure}{r}{0.5\textwidth}
\vspace{-0.2in}
  \begin{center}
    \includegraphics[width=0.498\textwidth]{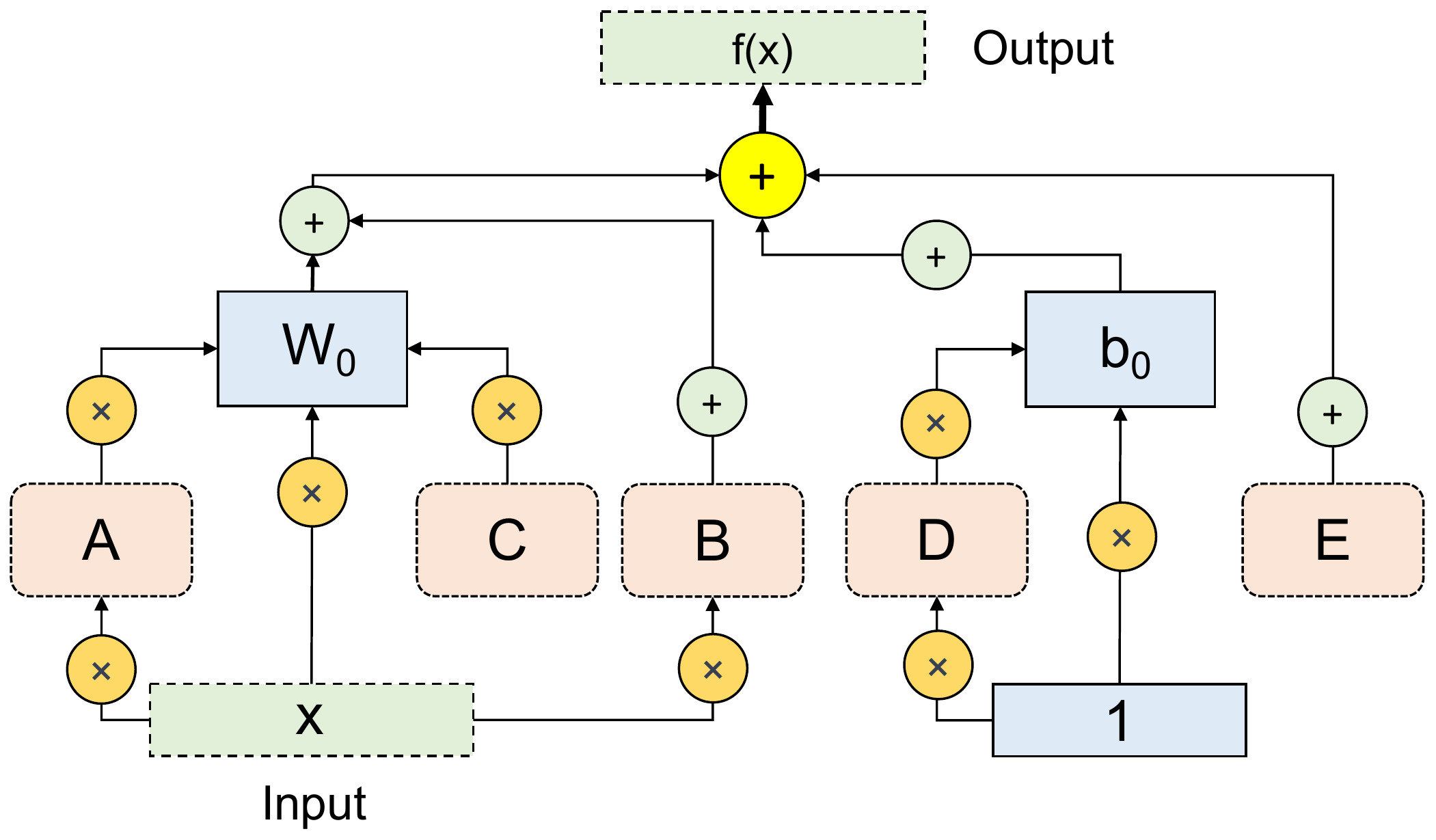}
  \end{center}
  \vspace{-0.15in}
  \caption{Schematic representation of a linear layer adapted with GLoRA.}
  \label{overview}
  \vspace{-0.1in}
\end{wrapfigure}
For model fine-tuning, we propose a unified formulation that encompasses tuning in both weight and feature space along with VPT-Deep level prompt design. Additionally, we adopt a re-parameterization strategy to incorporate auxiliary parameters into the adjacent projection weights during the inference stage. Broadly speaking, our method serves as a {\em superset} of all prior solutions, i.e., one-for-all mechanism. By setting different support tensors to zero, our GLoRA can be degraded to any of these predecessor methods. Unlike NOAH~\citep{zhang2022neural}, our architecture can be succinctly articulated as a unified mathematical equation. The consolidated formulation to represent all tunable spaces can be represented as follows:
\begin{equation} \label{eq:uni}
f(x)=\left(\mathbf{W}_{0}+\mathbf{W}_{0} \mathbf{A}+\mathbf{B}\right) x+\mathbf{C} \mathbf{W}_{0}+\mathbf{D} \mathbf{b}_{0}+\mathbf{E}+\mathbf{b}_{0} 
\end{equation}
where $\mathbf{A}$, $\mathbf{B}$, $\mathbf{C}$, $\mathbf{D}$, $\mathbf{E}$ are the trainable support tensors for downstream tasks in our GLoRA, $\mathbf{W}_{0}$ and $\mathbf{b}_{0}$ are frozen during whole fine-tuning. $\mathbf{A}$ is utilized to scale the weight. $\mathbf{B}$ has the role to scale the input and shift the weight. $\mathbf{C}$ is the layer-wise prompt serving a similar function of VPT-Deep, $\mathbf{D}$ and $\mathbf{E}$ are used to scale and shift the bias, respectively. A detailed illustration is shown in Figure~\ref{overview}.

\noindent{\bf Module Design}. 
In this subsection, we delineate the methodology for designing layer-wise adaptors or prompt modules for $\mathbf{A}$, $\mathbf{B}$, $\mathbf{C}$, $\mathbf{D}$, $\mathbf{E}$. In a broad sense, these can take the form of $\texttt{scalars}$, $\texttt{vectors}$, $\texttt{low-rank decompositions}$, or $\texttt{none}$. Based on the role of these trainable support tensors, they can be sampled from the following respective search spaces:
\begin{equation}
\begin{array}{ll}
\mathbf{A} = \{\mathrm{LoRA}, \mathrm{vector}, \mathrm{scalar}, \mathrm{none}\} \\ 
\mathbf{B} = \{\mathrm{LoRA}, \mathrm{vector}, \mathrm{scalar}, \mathrm{none}\} \\
\mathbf{C} = \{\mathrm{LoRA}, \mathrm{vector}, \mathrm{none} \} \\ 
\mathbf{D} = \{\mathrm{vector}, \mathrm{scalar}, \mathrm{none}\} \\ 
\mathbf{E} = \{\mathrm{vector}, \mathrm{scalar}, \mathrm{none}\} 
\end{array}
\label{support_tensor}
\end{equation}
where $\texttt{none}$ indicates zero, if all the trainable support tensors are zero, the model will be degraded to the original formulation and training recipe. In particular, suppose $\mathbf{W}_{0} \in \mathbb{R}^{d_{2} \times d_{1}}$ is the original weight matrix, with input and output channels denoted by $d_{1}$ and $d_{2}$ respectively. For every layer, we define $\mathbf{A}_{d} \in \mathbb{R}^{d_{2} \times r}$, $\mathbf{A}_{u} \in \mathbb{R}^{r \times d_{1}}$, $\mathbf{B}_{d} \in \mathbb{R}^{d_{2} \times r}$, $\mathbf{B}_{u} \in \mathbb{R}^{r \times d_{1}}$, $\mathbf{C}_{d} \in \mathbb{R}^{d_{2} \times r}$, $\mathbf{C}_{u} \in \mathbb{R}^{r \times 1}$, $\mathbf{D} \in \mathbb{R}^{d_{2} \times 1}$ and $\mathbf{E} \in \mathbb{R}^{d_{2} \times 1}$. We also define a multi-path supernet of all possible subnets and randomly sample a subnet during any given supernet training iteration for optimization. A subnet comprises of a single path network with different layerwise support tensors sampled from Eq. \ref{support_tensor}. Depending upon the current subnet configuration, in case of LoRA with rank $r_1 < r$, $\mathbf{A}_{d}^{r_1} \in \mathbb{R}^{d_{2} \times r_1}$, $\mathbf{A}_{u}^{r_1} \in \mathbb{R}^{r_1 \times d_{1}}$ is indexed from $\mathbf{A}_{d}$ and $\mathbf{A}_{u}$ respectively; and $\mathbf{A} = \mathbf{A}_{d}^{r_1} \times \mathbf{A}_{u}^{r_1}$ is used as the final tensor, in case of vector $\mathbf{A} \in \mathbb{R}^{d_{2} \times 1}$ is indexed from $\mathbf{A}_{d}$ and in case of scalar $\mathbf{A} \in \mathbb{R}^{1 \times 1}$ is indexed from $\mathbf{A}_{d}$. A similar strategy is followed for all other support tensors depending upon the current sampled configuration in the subnet. This weight entanglement strategy helps to increase the search space without increasing the number of parameters substantially and also shows faster convergence due to weight sharing in different subnets.

\subsection{Structural Re-parameterization Design and Inference Efficiency Analysis}

The fundamental factor enabling model re-parameterization~\citep{ding2021repvgg,hu2021lora} is the elimination of non-linearity amidst adjacent transformations, thereby permitting the absorption of supplementary parameters into the preceding ones. As mentioned in RepAdapter~\citep{luo2023towards}, the removal of such non-linear layers does not detrimentally impact the performance of the networks. The precise concept of GLoRA re-parameterization is explicated as follows:
\begin{equation}
f(x)=\mathbf{W}_\mathrm{uni}x+\mathbf{b}_\mathrm{uni}
\end{equation}
where $\mathbf{W}_\mathrm{uni}$ and $\mathbf{b}_\mathrm{uni}$ are our final unified trained weight and bias in GLoRA. They are re-parameterized according to Eq.~\ref{eq:uni}:
\begin{equation}
\mathbf{W}_\mathrm{uni}=\mathbf{W}_{0}+\mathbf{W}_{0} \mathbf{A}+\mathbf{B}
\end{equation}
\begin{equation}
\mathbf{b}_\mathrm{uni}=\mathbf{C} \mathbf{W}_{0}+\mathbf{D} \mathbf{b}_{0}+\mathbf{E}+\mathbf{b}_{0} 
\end{equation}
As a result, the re-parameterization strategy we employ, which integrates learnable parameters into the existing weight matrix offers a distinct advantage as it imposes no additional computational burden during the inference phase. This is further discussed in Section \ref{discussion} where we provide thorough inference efficiency analysis of GLoRA compared to exisitng works.

\subsection{Evolutionary Search for Optimal Layer-wise Configurations}

Our design for a unified adaptor is implemented on a per-layer basis, thus allowing for heterogeneity across different layers. To identify the optimal configuration for each layer, we employ the evolutionary search method~\citep{zhang2022neural, shen2021partial}, which offers a balance of efficiency and effectiveness. Although the training time may increase due to this search process, it is important to note that existing work~\citep{zhang2022neural} necessitate an extensive hyperparameter search (such as low-rank in LoRA and FacT, as well as position and size of adapter modules in Adapter~\citep{houlsby2019parameter}, dimension and structure configuration in RepAdapter~\citep{luo2023towards}, among others), as presented in Appendix. Our unified support tensor design conducts an implicit search that eliminates the need for manual hyperparameter tuning. Therefore, any augmentation in training time is reasonable and well-justified. More details regarding evolutionary search are in Appendix.

\subsection{GLoRA with Higher Capacity}

Model capacity refers to the capability of a model to approximate a diverse range of functions. A method for regulating the capacity of a learning algorithm involves selecting an appropriate hypothesis space, essentially a set of functions that the learning algorithm is permitted to consider as potential solutions. The Vapnik-Chervonenkis Dimension (VC Dimension)~\citep{vapnik2015uniform}, a measure of the capacity and complexity of a statistical algorithm, can be leveraged to provide a formal evidence for this assertion.

\begin{theorem} 
Suppose \(\mathbf d_\mathrm{vc}(\mathcal{H})\) is the VC dimension of any finite hypothesis \(\mathcal{H}\). If \(\mathcal{H}_\mathrm{i} \subseteq \mathcal{H}_\mathrm{uni}\),
\[ \mathbf d_\mathrm{vc}(\mathcal{H}_\mathrm{uni}) 	- \mathbf d_\mathrm{vc}(\mathcal{H}_\mathrm{i}) \geq \epsilon \ \ \ \ \  s.t. \ \ \ \epsilon \geq 0  \]
\label{higher_capacity}
\end{theorem}
\vspace{-0.25in}
In the context of GLoRA, $\mathcal{H}_\mathrm{i}$ denotes the hypothesis space of a randomly sampled subnet and $\mathcal{H}_\mathrm{uni}$ denotes the hypothesis space of the complete supernet. The validity of this theorem stems from the inherent property of our problem context, where the hypothesis space $\mathcal{H}_\mathrm{i}$ is a subset of $\mathcal{H}_\mathrm{uni}$ in our context. $\mathcal{H}_\mathrm{uni}$ encompasses all possible shattered scenarios of $\mathcal{H}_\mathrm{i}$. For the extreme case where the VC dimension $\mathbf d_\mathrm{vc}(\mathcal{H}_\mathrm{o})$ ($\mathcal{H}_\mathrm{o}$ is the 
difference set of $\mathcal{H}_\mathrm{uni}$ and $\mathcal{H}_\mathrm{i}$) is 0, 
the error $\epsilon$ will be zero. As per learning theory, a higher VC dimension implies greater model flexibility and capability of our approach. Clearly, Theorem \ref{higher_capacity} holds for GLoRA and thus it experiences a greater model capacity.

\section{Experiments} \label{exps}

\noindent{\bf Datasets.} 
We thoroughly evaluate GLoRA on VTAB-1K~\citep{zhai2020the} benchmark for various parameter budgets. VTAB-1K comprises 19 image classification tasks clustered into three domains: (i) Natural images; (ii) Specialized tasks consisting of remote sensing and medical datasets; and (iii) Structured tasks focusing on scene structure understanding. To examine the ability on few-shot learning, we evaluate GLoRA on five fine-grained visual recognition few-shot datasets: Food101 \citep{bossard14}, OxfordFlowers102 \citep{nilsback2006visual}, StandfordCars \citep{krause20133d}, OxfordPets \citep{parkhi2012cats}, and FGVCAircraft \citep{maji2013fine}. Following previous work \citep{jie2022fact}, we evaluate 1, 2, 4, 8, and 16-shot settings. Next, to show the domain generalization capabilities of GLoRA, we train it on ImageNet \citep{deng2009imagenet} for a 16-shot setting and test on four out-of-domain datasets including ImageNetV2 \citep{recht2019imagenet}, ImageNet-Sketch \citep{wang2019learning}, ImageNet-A \citep{hendrycks2021natural}, and ImageNet-R \citep{hendrycks2021many}. Finally, we show the performance of GLoRA on the Open LLM Leaderboard which consists of four datasets with varying prompt shots, namely AI2 Reasoning Challenge (25-shot) \citep{clark2018think}, TruthfulQA (0-shot) \citep{truthful}, HellaSwag (10-shot) \citep{zellers-etal-2019-hellaswag} and MMLU (5-shot) \citep{hendrycks2020measuring}.

\noindent{\bf Network Architecture and Implementation Details.} 
For all the vision experiments, we utilize ViT-B \citep{dosovitskiyimage}, a model pre-trained on ImageNet-21K, as our foundational model. 
For the language experiments, we consider two foundational base models: LLaMA-1-7B~\citep{touvron2023llama} and LLaMA-2-7B \citep{touvron2023llama2}.

Our supernets undergo a training process spanning 500 epochs and 15 epochs for vision and language datasets respectively, operating with a constant batch size of 64 and a cosine learning rate scheduler. It is crucial to highlight that this precise policy demonstrates robust efficacy across all settings, regardless of the dataset in use. Post the training of supernet, we perform an evolutionary search on the validation set to pinpoint the optimal task-specific subnet, finalized for implementation. Finally, we report the performance of the searched subnet on the test set. In Appendix, we provide further insights into dataset-specific learning rates and specific settings for different datasets.

\vspace{-0.05in}
\subsection{Results on VTAB-1K}

We train three different GLoRA supernet configurations to vary the number of trainable parameters. The difference among them is only the LoRA dimensions in the search space which varies {from} 8 and 4 in the largest model, 4 and 2 in the intermediate model, and 2 in the smallest model. This added parameter flexibility in our approach allows for user-defined trainable parameter count in the final models. Results on the VTAB-1K benchmark are shown in Table \ref{tab:vtab}. We push the state-of-the-art in parameter-efficient transfer learning by up to 2.9\%. Impressively, our smallest model already surpasses all existing approaches by a significant margin. It is worth noting that GLoRA performs competitively across datasets, in contrast, prior all existing works tend to fail on at least one, proving GLoRA's high generalization capabilities. GLoRA pushes the state of the art in as many as 14 out of 19 datasets under VTAB-1K while maintaining commendable performance on the others.

\begin{table*}[h]
\vspace{-0.1in}
\centering
\caption{\textbf{Full results on VTAB-1K benchmark}. ``\# params'' specifies the number of trainable parameters {in backbones}. Average accuracy and \# params are averaged over group-wise mean values.}
\label{tab:vtab}
\setlength{\tabcolsep}{0.3pt}
\resizebox{\textwidth}{!}{
\begin{tabular}{p{2.2cm}<{}p{0.9cm}<{\centering}|p{0.75cm}<{\centering}p{0.75cm}<{\centering}p{0.75cm}<{\centering}p{0.75cm}<{\centering}p{0.75cm}<{\centering}p{0.75cm}<{\centering}p{0.75cm}<{\centering}p{0.75cm}<{\centering}|p{0.75cm}<{\centering}p{0.75cm}<{\centering}p{0.75cm}<{\centering}p{0.75cm}<{\centering}|p{0.75cm}<{\centering}p{0.75cm}<{\centering}p{0.75cm}<{\centering}p{0.75cm}<{\centering}p{0.75cm}<{\centering}p{0.75cm}<{\centering}p{0.75cm}<{\centering}p{0.75cm}<{\centering}|p{0.75cm}<{\centering}}
\toprule[1.5pt]
\multicolumn{3}{c|}{}&\multicolumn{7}{c|}{\textbf{Natural}}&\multicolumn{4}{c|}{\textbf{Specialized}}&\multicolumn{8}{c|}{\textbf{Structured}}&\\
&\multicolumn{1}{c|}{{\rotatebox[origin=c]{90}{\# param (M)}}}
&\multicolumn{1}{c|}{{\rotatebox[origin=c]{90}{Inference Cost}}}
&\multicolumn{1}{c}{{\rotatebox[origin=c]{90}{Cifar100}}}
&\multicolumn{1}{c}{{\rotatebox[origin=c]{90}{Caltech101}}}
&\multicolumn{1}{c}{{\rotatebox[origin=c]{90}{DTD}}}
&\multicolumn{1}{c}{{\rotatebox[origin=c]{90}{Flower102}}}
&\multicolumn{1}{c}{{\rotatebox[origin=c]{90}{Pets}}}
&\multicolumn{1}{c}{{\rotatebox[origin=c]{90}{SVHN}}}
&\multicolumn{1}{c|}{{\rotatebox[origin=c]{90}{Sun397}}}
&\multicolumn{1}{c}{{\rotatebox[origin=c]{90}{Camelyon}}}
&\multicolumn{1}{c}{{\rotatebox[origin=c]{90}{EuroSAT}}}
&\multicolumn{1}{c}{{\rotatebox[origin=c]{90}{Resisc45}}}
&\multicolumn{1}{c|}{{\rotatebox[origin=c]{90}{Retinopathy}}}
&\multicolumn{1}{c}{{\rotatebox[origin=c]{90}{Clevr-Count}}}
&\multicolumn{1}{c}{{\rotatebox[origin=c]{90}{Clevr-Dist}}}
&\multicolumn{1}{c}{{\rotatebox[origin=c]{90}{DMLab}}}
&\multicolumn{1}{c}{{\rotatebox[origin=c]{90}{KITTI-Dist}}}
&\multicolumn{1}{c}{{\rotatebox[origin=c]{90}{dSpr-Loc}}}
&\multicolumn{1}{c}{{\rotatebox[origin=c]{90}{dSpr-Ori}}}
&\multicolumn{1}{c}{{\rotatebox[origin=c]{90}{sNORB-Azim}}}
&\multicolumn{1}{c|}{{\rotatebox[origin=c]{90}{sNORB-Ele}}}
&\multicolumn{1}{c}{{\rotatebox[origin=c]{90}{Average}}}\\
\specialrule{0em}{1pt}{1pt}
\hline
\specialrule{0em}{1pt}{1pt}
\multicolumn{23}{l}{\emph{Traditional Finetuning}}\\
\hline
\specialrule{0em}{1pt}{1pt}
Full&85.8&-&68.9&87.7&64.3&97.2&86.9&87.4&38.8&79.7&95.7&84.2&73.9&56.3&58.6&41.7&65.5&57.5&46.7&25.7&29.1&68.9 \\
Linear&0&-&64.4&85.0&63.2&97.0&86.3&36.6&51.0&78.5&87.5&68.5&74.0&34.3&30.6&33.2&55.4&12.5&20.0&9.6&19.2&57.6\\
\hline
\specialrule{0em}{1pt}{1pt}
\multicolumn{23}{l}{\emph{PEFT methods}}\\
\hline
\specialrule{0em}{1pt}{1pt}
BitFit &0.10&-&72.8&87.0&59.2&97.5&85.3&59.9&51.4&78.7&91.6&72.9&69.8&61.5&55.6&32.4&55.9&66.6&40.0&15.7&25.1&65.2\\
VPT-Shallow&0.06&$\uparrow$&77.7&86.9&62.6&97.5&87.3&74.5&51.2&78.2&92.0&75.6&72.9&50.5&58.6&40.5&67.1&68.7&36.1&20.2&34.1&67.8\\
VPT-Deep &0.53&$\uparrow$&\textbf{78.8}&90.8&65.8&98.0&88.3&78.1&49.6&81.8&96.1&83.4&68.4&68.5&60.0&46.5&72.8&73.6&47.9&32.9&37.8&72.0 \\
Adapter &0.16&$\uparrow$&69.2&90.1&68.0&98.8&89.9&82.8&54.3&84.0&94.9&81.9&75.5&80.9&65.3&48.6&78.3&74.8&48.5&29.9&41.6&73.9 \\
AdaptFormer &0.16&$\uparrow$&70.8&91.2&70.5&99.1&90.9&86.6&54.8&83.0&95.8&84.4&\textbf{76.3}&81.9&64.3&49.3&80.3&76.3&45.7&31.7&41.1&74.7 \\
LoRA &0.29&-&67.1&91.4&69.4&98.8&90.4&85.3&54.0&84.9&95.3&84.4&73.6&82.9&\textbf{69.2}&49.8&78.5&75.7&47.1&31.0&44.0&74.5 \\
NOAH &0.36&$\uparrow$&69.6&92.7&70.2&99.1&90.4&86.1&53.7&84.4&95.4&83.9&75.8&82.8&68.9&49.9&81.7&81.8&48.3&32.8&\textbf{44.2}&75.5\\
FacT &0.07&-&70.6&90.6&70.8&99.1&90.7&88.6&54.1&84.8&96.2&84.5&75.7&82.6&68.2&49.8&80.7&80.8&47.4&33.2&43.0&75.6\\
SSF &0.24&-& 69.0& 92.6& 75.1& 99.4& 91.8& 90.2& 52.9& 87.4& 95.9& 87.4& 75.5& 75.9& 62.3& 53.3& 80.6& 77.3& \textbf{54.9}& 29.5& 37.9&75.7\\
RepAdapter &0.22&-&72.4&91.6&71.0&99.2&91.4&90.7&55.1&85.3&95.9&84.6&75.9&82.3&68.0&50.4&79.9&80.4&49.2&38.6&41.0&76.1\\
\hline
\specialrule{0em}{1pt}{1pt}
\rowcolor{lightgray}\textbf{GLoRA}&0.86&-&76.4&\textbf{92.9}&74.6&99.6&\textbf{92.5}&\textbf{91.5}&\textbf{57.8}&87.3&\textbf{96.8}&88.0&76.0&\textbf{83.1}&67.3&\textbf{54.5}&\textbf{86.2}&\textbf{83.8}&52.9&\textbf{37.0}&41.4&\textbf{78.0}\\
\rowcolor{lightgray}\textbf{GLoRA}&0.44&-&76.5&92.3&75.2&99.6&92.3&91.2&57.5&87.3&96.7&88.1&76.1&80.6&67.2&53.4&84.5&83.5&52.8&35.2&40.8&77.6\\
\rowcolor{lightgray}\textbf{GLoRA}&0.29&-&76.1&92.7&\textbf{75.3}&\textbf{99.6}&92.4&90.5&57.2&\textbf{87.5}&96.7&\textbf{88.1}&76.1&81.0&66.2&52.4&84.9&81.8&53.3&33.3&39.8&77.3\\
\bottomrule[1.5pt]
\end{tabular}
}
\vspace{-0.06in}
\end{table*}

\subsection{Results on Large Language Models}

\begin{table}[h]
\vspace{-0.15in}
\centering
\caption{Performance of GLoRA on few-shot generative language tasks with LLMs as backbones.}
\label{tab:swin}
\resizebox{0.97\textwidth}{!}{
\begin{tabular}{@{}lc|cccc|c@{}}
\toprule
Model & Dataset & ARC (25-s) & HellaSwag (10-s) & MMLU (5-s) & TruthfulQA (0-s) & Average \\ \midrule
LLaMA-1-7B & - &51.0 & 77.8 & 35.7 & 34.3 & 49.7\\
LoRA & Alpaca & 53.5 & 77.3 & 33.8 & 34.8 & 49.8 \\
\rowcolor{lightgray}GLoRA & Alpaca & 52.9 & 78.1 & 34.5 & 37.8 & 50.8  \\
LoRA & ShareGPT & 51.7 & 77.9 & 36.1 & 39.2 & 51.2 \\
\rowcolor{lightgray}GLoRA & ShareGPT & 53.2 & 77.4 & 36.2 & 43.9 & 52.7  \\ \hline
LLaMA-2-7B & - & 53.1 & 78.5 & 46.9 & 38.8 & 54.3\\
\rowcolor{lightgray}GLoRA & ShareGPT & 53.7 & 78.5 & 46.5 & 45.1 & 56.1\\
\bottomrule
\end{tabular}
}
\end{table}
We apply GLoRA for LLMs by solely tuning the attention layers. This contrasts with vision tasks where all linear layers are adapted, to maintain a fair comparison with vanilla LoRA. 
We start from the publicly available LLaMA-1-7B \citep{touvron2023llama} and LLaMA-2-7B \citep{touvron2023llama2} models and finetune them on the Alpaca \citep{alpaca} and ShareGPT dataset with only GLoRA support tensors trainable. For the evolutionary search, we use 5\% random data sampled from the 4 given datasets for model validation during the evolutions. We finally report the searched model's performance on 
the standard Open LLM Leaderboard\footnote{\url{https://huggingface.co/spaces/HuggingFaceH4/open_llm_leaderboard}}. GLoRA consistently outperforms the pre-trained LLM and the corresponding LoRA fine-tuned variants. We maintain consistent hyperparameters between LoRA and GLoRA for a fair comparison, more details are in the Appendix.

\subsection{Few-shot Learning}
\begin{figure}
  \centering
  \includegraphics[width=\textwidth]{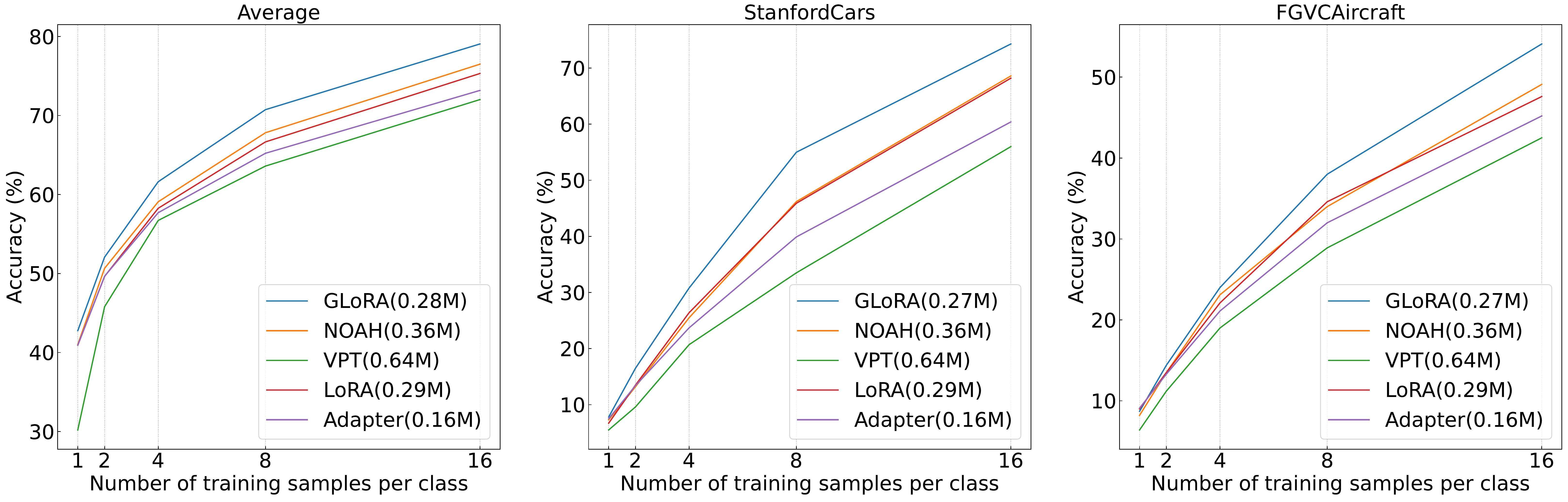}
  \vspace{-0.23in}
  \caption{\textbf{Results on few-shot learning datasets}. The baseline methods include Adapter, LoRA, VPT, NOAH. GLoRA consistently performs better across five datasets and a varying number of training examples per class. More comparisons are provided in Appendix~\ref{appendix_few_shot}.}
  \label{fs_main}
  \vspace{-0.15in}
\end{figure}
To extend the evaluation of GLoRA under conditions of limited data availability, we present the performance of GLoRA on fine-grained visual recognition datasets as the few-show learning, comparing it with LoRA, Adapter, VPT, and NOAH. The results at 1, 2, 4, 8, and 16 shots are illustrated in Figure~\ref{fs_main} and Figure \ref{fs} of Appendix. GLoRA demonstrates superior performance across the majority of the few-shot learning datasets, consistently outperforming the performance of existing methods by a large margin with similar parameter counts. Interestingly, on the Flowers102 dataset, all methods yield similar accuracy levels, attributable to the already exceptional overall performance. On the Food101 dataset, the average accuracy of GLoRA is on par with NOAH. From the first plot, we can observe that the average performance boost becomes more pronounced at higher shot scenarios, nevertheless, even at lower shot settings, the gains of our approach remain significant.

\subsection{Domain Generalization}
\begin{table}[h]
\centering
    \caption{\textbf{Results on domain generalization}. GLoRA is significantly better than the existing works.}
    \label{tab:dg}
    \resizebox{0.65\textwidth}{!}{
    \begin{tabular}{l ccccc}
    \toprule
    & \textbf{Source} & \multicolumn{4}{c}{\textbf{Target}} \\ \cmidrule(lr){2-2} \cmidrule(lr){3-6}
    & ImageNet & -Sketch & -V2 & -A & -R \\
    \midrule
    Adapter \cite{houlsby2019parameter} &70.5 & 16.4 & 59.1 & 5.5 & 22.1 \\
    VPT \cite{jia2022visual} & 70.5 & 18.3 & 58.0 &  4.6 & 23.2 \\
    LoRA \cite{hu2021lora} & 70.8 & 20.0 & 59.3 &  6.9 & 23.3 \\
    NOAH \cite{zhang2022neural} & 71.5 & 24.8 & 66.1 &  11.9 & 28.5  \\
  \rowcolor{lightgray}  GLoRA (0.29M) & 78.3 & 30.6 & 67.5 &  13.3 & 31.0 \\
    \bottomrule
    \end{tabular}
    }
    \vspace{-0.15in}
\end{table}
The capacity of out-of-domain generalization holds significant value for large-scale neural networks \citep{zhou2021domain}. Models fine-tuned via PEFT methods should exhibit enhanced domain generalization aptitude, thereby making them more applicable in real-world scenarios. We demonstrate the out-of-domain generalization capabilities of GLoRA in Table \ref{tab:dg}, where a single ImageNet-1K \citep{deng2009imagenet} fine-tuned GLoRA model is subjected to testing on out-of-domain datasets. Aligning with preceding research, we limit the number of training examples per class to 16 for this experiment. It is noteworthy that the performance for the fully-scaled ImageNet-1K fine-tuned model stands at 83.97\% \citep{dosovitskiyimage}, and our approach manages to narrow this performance gap, even within a 16-shot setting (78.3\%), thereby exhibiting superior few-shot learning on ImageNet-level datasets. Furthermore, the out-of-domain performance also witnesses a substantial boost in comparison to existing methods. When compared with LoRA, GLoRA enhances out-of-domain performance by as much as 100\% (ImageNet-A) and 50\% (ImageNet-Sketch).

\vspace{-0.1in}
\section{Analysis and Discussion}
\label{discussion}

\noindent{\bf Computational Cost.}
\begin{table}[]
\centering
\caption{Inference efficiency comparison of GLoRA with existing methods.}
\label{tab:inference}
\resizebox{0.8\textwidth}{!}{
\begin{tabular}{@{}cccccc@{}}
\toprule
\multirow{2}{*}{Method} & \multirow{2}{*}{$\uparrow$ \#Param(M)} & \multirow{2}{*}{$\uparrow$ FLOPs(G)} & \multicolumn{3}{c}{Throughput (imgs/sec)} \\ \cmidrule(l){4-6} 
                        &                               &                             & bs = 1      & bs = 4     & bs = 16     \\ \midrule
Full tuning             & 0                             & 0                           & 91.5        & 375.7      & 539.5       \\ \midrule
VPT \cite{jia2022visual}              & 0.55                          & 5.60                        & 86.1        & 283.5      & 381.5       \\
Adapter \cite{houlsby2019parameter}          & 0.16                          & 0.03                        & 70.9        & 306.6      & 504.7       \\
AdaptFormer \cite{chen2022adaptformer}           & 0.16                          & 0.03                        & 71.4        & 309.9      & 508.1       \\
NOAH \cite{zhang2022neural}                & 0.12                          & 0.02                        & 72.1        & 312.7      & 492.9       \\ \midrule
LoRA \cite{hu2021lora}                    & \multirow{2}{*}{0}                             & \multirow{2}{*}{0}                           & \multirow{2}{*}{91.5}        & \multirow{2}{*}{375.7}      & \multirow{2}{*}{539.6}       \\
$\!\!\!\!$GLoRA                   &                              &                            &        &       &        \\ \bottomrule
\end{tabular}
}
\vspace{-0.1in}
\end{table}
We show the final inference throughput of various PEFT methods in Table \ref{tab:inference}, computed on an NVIDIA 3090 GPU. The results highlight that GLoRA surpasses other competitive methods in performance, as it does not require any extra parameters or FLOPs during the inference stage. An additional advantage is its quicker adaptability in real-world scenarios, especially when prior or foundational models are already deployed.  
The weights of GLoRA can be directly loaded without necessitating any manual system modifications. As previously mentioned, GLoRA supports VPT-Deep level prompts via the support tensor $\mathbf{C}$, however, it does not impose any computational overhead due to its completely structural re-parameterization design.

\begin{wrapfigure}{r}{0.5\textwidth}
\vspace{-0.15in}
  \centering
  \includegraphics[width=0.5\textwidth]{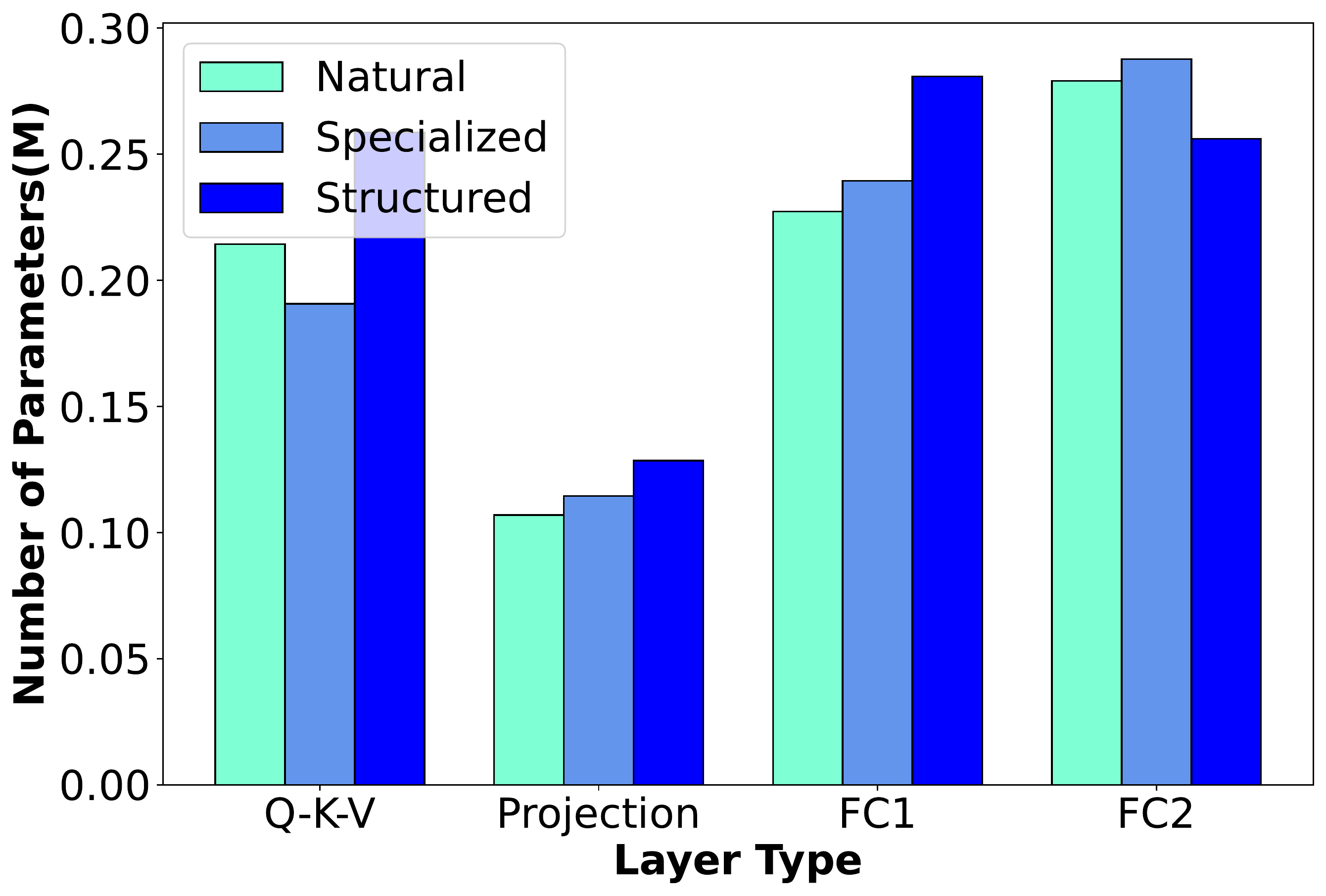}
  \vspace{-0.3in}
  \caption{Distribution of GLoRA (0.86M) parameters across layer types on VTAB-1K. {\em Q-K-V} and {\em Projection} are linear layers in MHSA module and {\em FC1} and {\em FC2} are linear layers in MLP module.}
  \vspace{-0.1in}
  \label{layer_types}
\end{wrapfigure}

\begin{figure}
  \centering
  \includegraphics[width=0.98\textwidth]{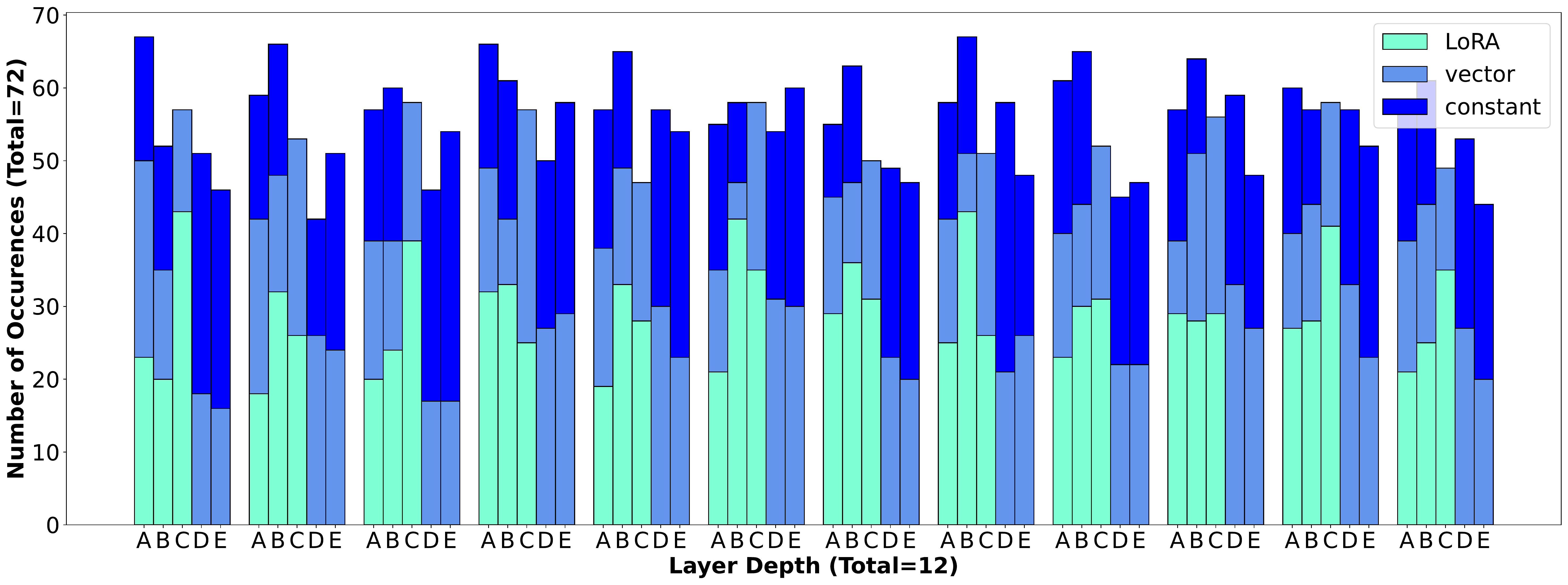}
  \vspace{-0.2in}
  \caption{Layerwise configuration of support tensors in GLoRA (0.86M) on VTAB-1K dataset. }
  \label{depth_wise}
  \vspace{-0.1in}
\end{figure}

\noindent{\bf Visualizations of searched fine-tuning strategy for each layer.} 
Figure \ref{layer_types} visually shows the distribution of trainable parameters across the four types of linear layers embodied in ViT-B. Notably, the projection layer possesses the minimum quantity of trainable parameters spanning across VTAB-1K categories. Generally, the MLP module hosts a substantially higher number of parameters compared to the MHSA. As anticipated, the structured group necessitates a greater number of parameters for adaptation due to a pronounced domain shift relative to ImageNet-1K \citep{deng2009imagenet}. Figure \ref{depth_wise} illustrates the layer-wise configuration of the support tensors as searched by the GLoRA algorithm. Each support tensor at every layer can potentially undergo 72 distinct adaptations across datasets. Support tensors $\mathbf{D}$ and $\mathbf{E}$ exhibit relatively low adaptation due to the prevalence of \texttt{none} adaptations, whereas $\mathbf{A}$ and $\mathbf{B}$ demonstrate a higher number of adaptations, though without a distinguishable pattern regarding the type of adaptation. It is important to underscore that even a basic scalar can function effectively as a support tensor, enabling GLoRA to maintain superior parameter efficiency despite adapting every linear layer.

\vspace{-0.1in}
\section{Related Work}
\vspace{-0.05in}
Given the rapid expansion in model size, numerous methods for parameter-efficient fine-tuning (PEFT) have been introduced in the field of NLP to streamline the optimization of large language models (LLMs)~\citep{DBLP:journals/corr/abs-2110-07602, zhangadaptive, hulora, liu2021gpt, li-liang-2021-prefix, lester2021power, zaken2022bitfit, houlsby2019parameter}. The effectiveness of parameter-efficient fine-tuning has been proven in a wide range of natural language processing tasks \citep{fu2022effectiveness,he2021towards}. In the vision domain, with the advent growth in the size of vision models \citep{dehghani2023scaling, kolesnikov2020big}, methods specifically focused on image modality have also been put forward \citep{jie2022fact, lian2022scaling, chen2022adaptformer, luo2023towards, zhang2022neural, jia2022visual, he2023sensitivityaware}. Among these methods, LoRA \citep{hulora} has proven to transfer well across modalities and tasks. This is partly due to the simplistic design strategy of LoRA which directly works over weight tensors, irrespective of model type or configuration. Additionally, unlike Adapters \citep{houlsby2019parameter, chen2022adaptformer} and Prompt tuning~\citep{jia2022visual}, LoRA does not add any additional inference parameters or latency due to structural re-parameterization (SR) design. RepAdapter \citep{luo2023towards} and SSF \citep{lian2022scaling} also propose an SR design for PEFT. However, RepAdapter is specific to model architectures and required manual designing for different layer configurations. SSF provides a simple baseline but suffers from low flexibility and capability due to adaptation limited in the activation space. FacT \citep{jie2022fact} further decomposes LoRA matrices for better parameter efficiency, but we argue that $<$1M parameter scale is fairly efficient for fine-tuning on a single GPU. Thus, due to the advantages of LoRA over other related works, it is of importance to increase the flexibility, scalability and adaptability of LoRA.

\vspace{-0.1in}
\section{Conclusion}
\vspace{-0.05in}

We have presented GLoRA, a generalized parameter-efficient fine-tuning approach that has successfully demonstrated its effectiveness and adaptability in enhancing the fine-tuning and transfer learning ability for the large-scale pre-trained models. By adopting a generalized low-rank adaptation and re-parameterization framework, GLoRA significantly reduces the number of parameters and computation required for fine-tuning, making it a more resource-efficient and practical method for real-world applications. 
The experiments conducted on a diverse range of tasks and datasets have substantiated the superiority of GLoRA over existing PEFT techniques, showcasing its scalability and adaptability. Moreover, the ablation studies have provided valuable insights into the inner workings and the relative importance of different GLoRA components. 
This work not only contributes to the improvement of the fine-tuning process for large-scale pre-trained vision or language models but also opens up new avenues for future work, including further exploration of generalized low-rank adaptation techniques, the development of hybrid approaches, and the refinement of search and optimization algorithms. These areas of research may continue to expand the accessibility and efficiency of transfer learning across a broader range of applications.

\bibliography{iclr2024_conference}
\bibliographystyle{iclr2024_conference}

\newpage

\appendix
\section*{\Large{Appendix}}
\section{Hyperparameters} \label{appen_hy}

\begin{table}[h]
\vspace{-0.2in}
\caption{Learning rate of dataset-specific supernet training on VTAB-1K datastet.}
\label{tab:lr_grid}
\resizebox{\textwidth}{!}{
\begin{tabular}{@{}lccccccccccccccccccc@{}}
\toprule
Dataset &  \rotatebox{90}{Cifar100} & \rotatebox{90}{Caltech101} & \rotatebox{90}{DTD} & \rotatebox{90}{Flowers102} & \rotatebox{90}{Pets} & \rotatebox{90}{SVHN} & \rotatebox{90}{Sun397} & \rotatebox{90}{Camelyon} & \rotatebox{90}{EuroSAT} & \rotatebox{90}{Resisc45} & \rotatebox{90}{Retinopathy} & \rotatebox{90}{Clevr-Count} & \rotatebox{90}{Clevr-Dist} & \rotatebox{90}{DMLab} & \rotatebox{90}{KITTI-Dist} & \rotatebox{90}{dSpr-Loc} & \rotatebox{90}{dSpr-Ori} & \rotatebox{90}{sNORB-Azim} & \rotatebox{90}{sNORB-Ele} \\ \midrule
LR & $5e^{-4}$ & $5e^{-4}$ & $5e^{-4}$ & $5e^{-4}$ & $5e^{-4}$ & $5e^{-4}$ & $5e^{-4}$ & $5e^{-4}$ & $5e^{-4}$ & $5e^{-4}$ & $1e^{-4}$ & $1e^{-4}$ & $1e^{-4}$ & $5e^{-4}$ & $5e^{-4}$ & $5e^{-4}$ & $5e^{-4}$ & $5e^{-4}$ & $1e^{-4}$ \\ \bottomrule
\end{tabular}
}
\vspace{-0.1in}
\end{table}

Our approach necessitates minimal adjustments to hyperparameters, with optimizer hyperparameters being the sole exception, thanks to the inherent search mechanism. Following prior studies \citep{dehghani2023scaling, chen2022adaptformer, zhang2022neural}, we employ the AdamW optimizer \citep{loshchilovdecoupled} for all our experiments. 

For the hyperparameter search in vision tasks, we primarily concentrate on the exploration of the learning rate for supernet training, limiting our search scope to two potential alternatives: $1e^{-4}$ and $5e^{-4}$. For a detailed setting of dataset-specific learning rates, please refer to Table \ref{tab:lr_grid}. All other training configurations strictly adhere to the exact training policy delineated in the works of \citep{jie2022fact, luo2023towards}. 
In the case of few-shot learning datasets and ImageNet, we use learning rates of $5e^{-4}$ and $1e^{-4}$ respectively, as the few-shot learning datasets are smaller if compared to 16-shot ImageNet dataset. 

For language modeling experiments, we use a learning rate of $2e^{-5}$ with cosine annealing and an equivalent batch size of 32 (using gradient accumulation) for both LoRA and GLoRA. Consequently, LoRA is trained for 3 epochs, and due to the supernet structure of GLoRA, we train it for 15 epochs. This is in line with vision experiments where LoRA is trained for 100 epochs and GLoRA supernet for 500 epochs. We justify these extra training epochs due to the fact that LoRA requires hyperparameter tuning (dropout rate, adaptation layer choice, alpha, etc.) while GLoRA, being a searched-based method, requires no such manual tuning. We provide more details, such as method-specific hyperparameters, in Appendix \ref{hparam} to justify GLoRA's extra training time.

\section{Evolutionary Search}

Evolutionary search consists of reproduction, crossover, and mutation stages. In our scenario, first, a population of support tensor strategies is embedded in vectors and initialized randomly. Each individual strategy consists of a description of a single subnet. After supernet training, we start to evaluate each individual subnet to obtain its accuracy on the validation set. Among these evaluated subnets we select the top $K$ as parents to produce posterity subnets. The next generation subnets are made by mutation and crossover stages. By repeating this process in iterations, we can find the best parameter-efficient fine-tuned subnet with the best validation performance. 

We first randomly sample 50 subnets from the supernet and then perform an evolutionary search for 20 and 5 epochs on vision and language tasks, respectively. Each step of random pick / crossover / mutation produces 50 new subnets. The probability for crossover and mutation is set to 0.2. Note that we did not perform any hyperparameter search over the evolution hyperparameters, and hence the performance might even improve after tuning the evolution hyperparameters.

\section{Hierarchical Transformer}

We show the performance of GLoRA on the Swin-B backbone in Table \ref{tab:swin}. We follow a dataset-specific learning rate searching similar to ViT-B and also add GLoRA to the reduction linear layer in Swin architecture to maintain uniformity and avoid architecture-specific tuning. GLoRA can adapt to any layer irrespective of architecture configuration and perform well across tasks and datasets which can be clearly seen in Table \ref{tab:swin}, where GLoRA outperforms all existing works by a fair margin.
\begin{table}[htp]
\centering
\caption{Performance on VTAB-1K benchmark with Swin-B model pre-trained on ImageNet-21K as the backbone. }
\label{tab:swin}
\begin{tabular}{@{}ccccc@{}}
\toprule
Method & Natural & Specialized & Structured & Average \\ \midrule
Full   &  79.2 & 86.2 & 59.7 & 75.0      \\
Linear &   73.5 & 80.8 & 33.5 & 62.6    \\
BitFit    &  74.2        &  80.1            &  42.4          &  65.6       \\
VPT    &  76.8       &   84.5          &   53.4         &  71.6       \\
FacT   &  82.7 & 87.5 & 62.0          &  77.4       \\
RepAdapter   &  83.1       &  86.9           &  62.1          &  77.4       \\
GLoRA  &  83.7       & 88.7            &  61.9          &   78.1      \\ \bottomrule
\end{tabular}
\end{table}

\section{Training Time}
\label{hparam}
\begin{table}[]
\caption{Manual design choices in existing works.}
\label{tab:design_choice}
\begin{tabular}{@{}cc@{}}
\toprule
Method      & Design Choices/Hyperparameters                                               \\ \midrule
VPT         & Prompt Length, Prompt Location, Prompt Depth                           \\
AdaptFormer & Adapter Location, Scaling Factor, Hidden dimension, Insertion Form     \\
NOAH       & VPT choices, Adapter choices, LoRA rank                                     \\ 
RepAdapter  & Adapter Location, Number of groups, Hidden dimension, Adapter variants \\
FacT        & Decomposition method, Scaling factor, Decomposition Rank               \\
GLoRA       & LoRA ranks in search space                                             \\ \bottomrule
\end{tabular}
\end{table}
Our GLoRA, being a search-based approach for PEFT, naturally incurs increased training time due to the requirements of supernet training and evolutionary search. However, it is critical to underscore that all current methods necessitate a manual search for design choices, as evidenced in Table \ref{tab:design_choice}. This necessity significantly inflates the total training time for a specific dataset, due to the broad search within these design choices. GLoRA streamlines this process through an automated evolutionary search mechanism, thus leveraging the benefit of an expansive search space.

\section{Search Space}
In this section, we analyze the computation of the possible number of subnets within our GLoRA-adapted supernet. Each layer offers $4, 4, 3, 3$, and $3$ options for the support tensor $\mathbf{A}$, $\mathbf{B}$, $\mathbf{C}$, $\mathbf{D}$, and $\mathbf{E}$, respectively. This results in $432$ possible configurations for a single linear layer. In our implementation, we incorporate $48$ such layers within ViT-B, yielding a total of $432 \times 48 = 20,736$ subnets being explored within GLoRA. This figure can escalate if multiple LoRA ranks coexist within the same search space. For instance, we allow ranks 8 and 4 in our largest GLoRA models, leading to $82,944$ distinct subnets. Furthermore, owing to the phenomenon of weight entanglement as per \citep{AutoFormer}, comparable performance is maintained across all subnets, even if they are not all explored during the training of the supernet.

\section{Support Tensor}

In this section, we justify the choices of support tensors in our framework. Consider a linear layer that facilitates the transformation of inputs from a $d_1$ dimensional space to a $d_2$ dimensional space, with a corresponding weight matrix $\mathbf{W}_{0} \in \mathbb{R}^{d_2 \times d_1}$. Given that $\mathbf{A}$ is tasked with scaling $\mathbf{W}_0$, $\mathbf{A}$ could feasibly belong to $\mathbb{R}^{d_2 \times d_1}$, $\mathbb{R}^{d_2 \times 1}$, or $\mathbb{R}^{1 \times 1}$. These matrix dimensions are respectively indicative of LoRA, vector, and scalar operations. It is pertinent to note that in scenarios where $\mathbf{A} \in \mathbb{R}^{d_2 \times d_1}$, LoRA is realized via corresponding matrices $\mathbf{A}_{d} \in \mathbb{R}^{d_2 \times r}$ and $\mathbf{A}_{u} \in \mathbb{R}^{r \times d_1}$. A parallel scrutiny of other support tensors would result in determining the appropriate support tensor choice, as elaborated in Section~\ref{method} of the main paper.

\section{Fine-tuned Embedding Visualization}
We present feature visualizations of the ViT-B model adapted via GLoRA and FacT \citep{jie2022fact} methods applied to the SVHN dataset. We select FacT as opposed to LoRA \citep{hulora}, given that FacT constitutes a direct mathematical enhancement over LoRA and presently represents the state-of-the-art. A clear distinction can be discerned whereby GLoRA exhibits superiorly segregated clusters in comparison to FacT. Further, the delineations are broader, and the clusters demonstrate a higher degree of concentration, signaling the heightened discriminative capacity of the GLoRA-adapted model features.

\begin{figure}[h]
  \centering
  \includegraphics[width=0.9\textwidth]{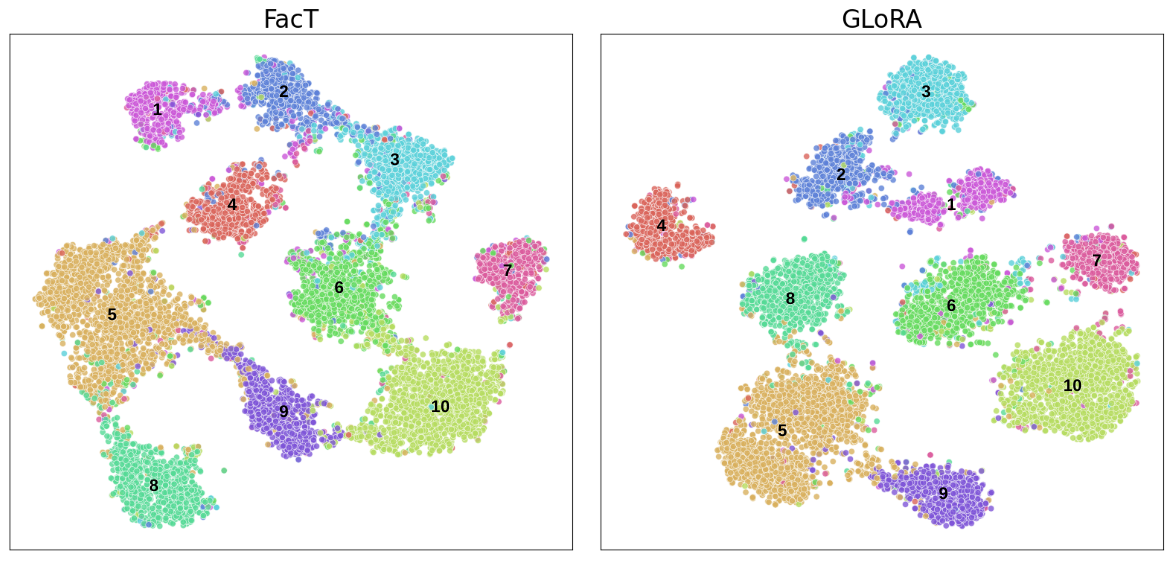}
  \vspace{-0.2in}
  \caption{Visualization of features from SVHN dataset by t-SNE \citep{van2008visualizing}.}
  \label{fs}
\end{figure}

\section{More Results on Few-shot Learning Datasets} \label{appendix_few_shot}

As shown in~\ref{fs}, the baseline methods include Adapter, LoRA, VPT, NOAH. GLoRA consistently performs better across five datasets and a varying number of training examples per class.

\begin{figure}[h]
  \centering
  \includegraphics[width=0.95\textwidth]{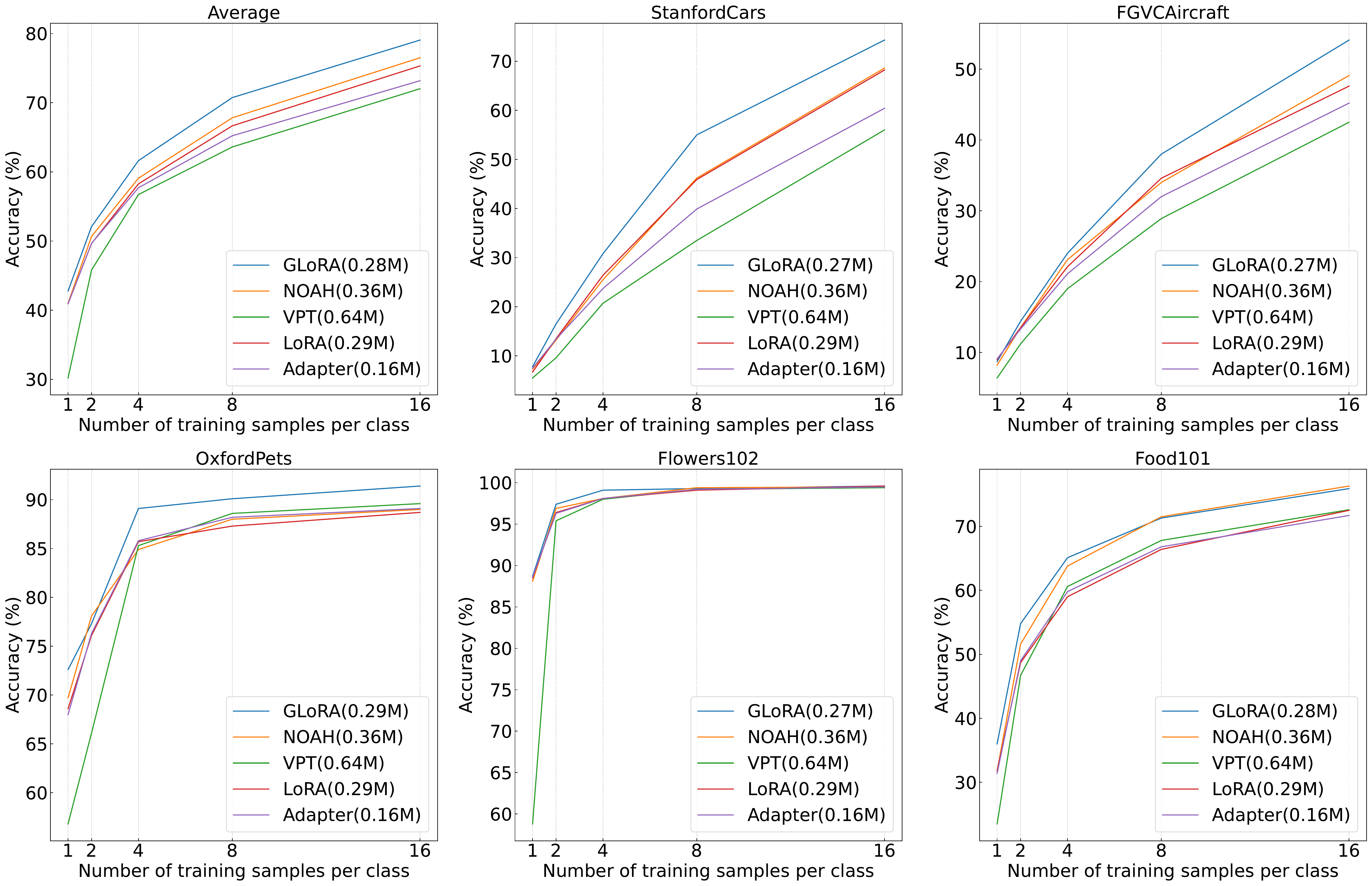}
  \vspace{-0.15in}
  \caption{{More results on few-shot learning datasets}.}
  \label{fs}
\end{figure}

\end{document}